\documentclass[letterpaper, 10 pt, conference]{ieeeconf}  

\IEEEoverridecommandlockouts                              

\usepackage{graphics} 
\usepackage{graphicx}
\usepackage{epsfig} 
\usepackage{times} 
\usepackage{amsmath} 
\usepackage{amssymb}  
\usepackage{booktabs}
\usepackage{tabularx}
\usepackage{cite}
\usepackage{balance}

\title{\LARGE \bf
COVLM-RL: Critical Object-Oriented Reasoning for Autonomous Driving Using VLM-Guided Reinforcement Learning }

\author{Lin Li$^1$, Yuxin Cai$^1$, Jianwu Fang$^2$, Jianru Xue$^2$, Chen Lv$^{1}$$^*$
\thanks{$^1$Lin Li, Yuxin Cai, and Chen Lv are with the School of Mechanical and Aerospace Engineering, Nanyang Technological University, 639798, Singapore. $^*$ Corresponding Author: Chen Lv
        ({\small lyuchen@ntu.edu.sg}).}%
\thanks{$^{2}$Jianwu Fang and Jianru Xue are with the National Key Laboratory of Human-Machine Hybrid Augmented Intelligence, National Engineering Research Center for Visual Information and Applications, Institute of Artificial Intelligence and Robotics, Xi'an Jiaotong University, Xi'an 710049, China.}
\thanks{This work was supported in part by the Agency for Science, Technology and Research (A*STAR), Singapore, through the MTC Individual Research Grant under Grant M22K2c0079; the Ministry of Education (MOE), Singapore, through the Tier 2 Grant under Grant MOE-T2EP50222-0002; the Desay SV Automotive Singapore Private Ltd., under Project NTU REF 2018-0980, and the National Key Research and Development Program of China (2024YFE0210700).}%
}
\begin{document}

\maketitle
\thispagestyle{empty}
\pagestyle{empty}

\begin{abstract}
End-to-end autonomous driving frameworks face persistent challenges in generalization, training efficiency, and interpretability. 
While recent methods leverage Vision-Language Models (VLMs) through supervised learning on large-scale datasets to improve reasoning, they often lack robustness in novel scenarios. Conversely, reinforcement learning (RL)-based approaches enhance adaptability but remain data-inefficient and lack transparent decision-making. 
To address these limitations, we propose \textbf{COVLM-RL}, a novel end-to-end driving framework that integrates Critical Object-oriented (CO) reasoning with VLM-guided RL. 
Specifically, we design a Chain-of-Thought (CoT) prompting strategy that enables the VLM to reason over critical traffic elements and generate high-level semantic decisions, effectively transforming multi-view visual inputs into structured semantic decision priors. These priors reduce the input dimensionality and inject task-relevant knowledge into the RL loop, accelerating training and improving policy interpretability. However, bridging high-level semantic guidance with continuous low-level control remains non-trivial. 
To this end, we introduce a consistency loss that encourages alignment between the VLM's semantic plans and the RL agent's control outputs, enhancing interpretability and training stability. 
Experiments conducted in the CARLA simulator demonstrate that COVLM-RL significantly improves the success rate by 30\% in trained driving environments and by 50\% in previously unseen environments, highlighting its strong generalization capability.
\end{abstract}

\section{INTRODUCTION}

End-to-end decision-making frameworks have emerged as a promising alternative to traditional modular pipelines in autonomous driving, aiming to address long-standing issues such as poor scenario generalization, error accumulation across modules, and system complexity \cite{10919979, sae5, 10208229, sae6}. These frameworks learn a direct mapping from sensory inputs to control actions, bypassing the need for manually designed perception, planning, and control modules \cite{dong2024generalizing, sae1, sae2, prakash2021multi, sae3, sae4}. Input modalities typically include camera images, LiDAR point clouds, and other on-board sensor data. While LiDAR offers precise spatial information, its high cost and computational demands limit its adoption in decision-making systems. As a result, image-based models have become the mainstream in recent end-to-end driving research.

Within the vision-based end-to-end paradigm, RL offers greater potential for generalization compared to supervised learning (SL) and imitation learning (IL). While SL and IL approaches have shown success by leveraging large-scale datasets of paired observations and expert control actions \cite{8569992}, their performance is fundamentally constrained by the diversity and quality of the training data. Collecting such data in real-world driving environments is costly, time-consuming, and often unsafe. In contrast, RL enables vehicles to learn policies through trial-and-error interaction, allowing them to adapt to novel scenarios and reducing reliance on curated datasets. Motivated by these advantages, this work adopts RL to train an end-to-end driving policy in CARLA \cite{ Dosovitskiy17}, a high-fidelity autonomous driving simulator that supports safe and scalable exploration of diverse traffic scenarios and visual conditions.


\begin{figure}
    \centering
    \includegraphics[width=\linewidth]{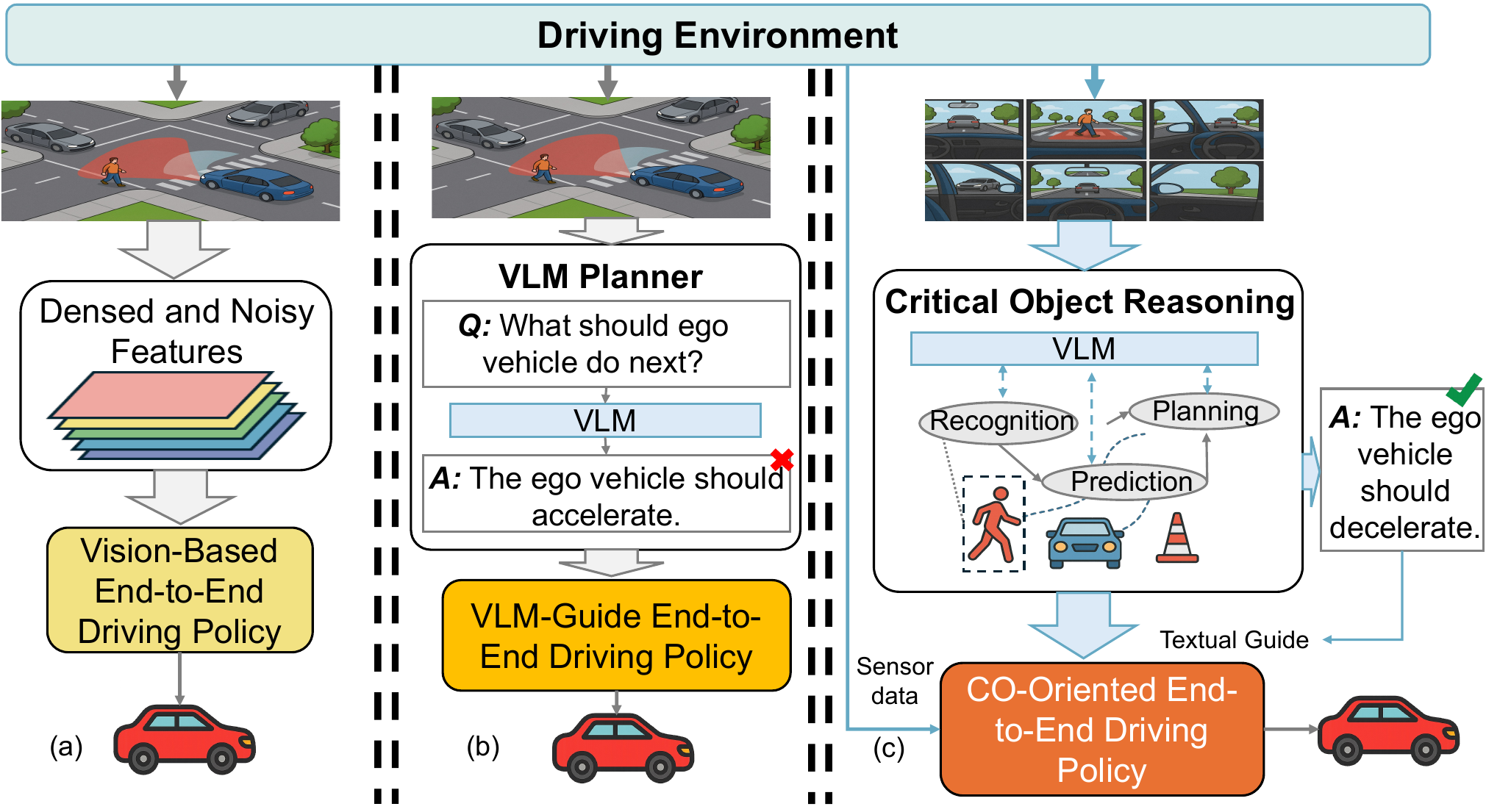}
    \caption{Comparison of existing End-to-End driving policies: (a)Vision-based End-to-End driving policy. (b) VLM-guided End-to-End driving policy. 
    (c) Our proposed COVLM-RL, a novel end-to-end driving framework that integrates critical object-oriented reasoning with VLM-guided RL.
    }
    \label{fig:problem}
\end{figure}
While RL offers strong generalization potential, existing vision-based RL approaches remain limited by low sample efficiency and poor interpretability. These models typically rely on standard convolutional visual encoders, such as CNNs or ResNet50 \cite{9164410}, to extract features from raw images, as shown in Fig.~\ref{fig:problem}(a). However, the resulting representations are often high-dimensional and redundant, which increases the policy's exploration space, slows convergence, and hinders learning efficiency. In addition, policies trained in fixed environments often rely on memorizing latent patterns in the visual input, rather than developing structured, semantic reasoning. As a result, their performance degrades significantly in novel scenarios. 
Most critically, end-to-end RL models operate as black boxes, offering little transparency into how control decisions are made-an issue that raises serious safety and reliability concerns in real-world autonomous driving. 
These challenges highlight the need for decision-making frameworks that incorporate high-level reasoning and structured semantic understanding to support both efficient learning and interpretable behavior.

Recently, rapid advancement of foundation models-such as large language models (LLMs) and vision-language models (VLMs)-has opened new avenues for improving RL in embodied decision-making tasks \cite{zhang2024vision,li2025benchmark}. These models offer rich semantic understanding and reasoning capabilities that can help abstract raw sensory inputs into interpretable, high-level representations. Some recent works have explored using pre-trained VLMs to assist RL agents, as shown in Fig.~\ref{fig:problem}(b) often by extracting visual features or generating goal-relevant language descriptions. However, these approaches typically treat VLMs as passive feature extractors or static priors, without fully leveraging their reasoning capabilities. Moreover, they often lack mechanisms to align high-level semantic outputs with low-level control actions, which limits their effectiveness in dynamic decision-making contexts.

To address these limitations, we propose a novel VLM-guided RL framework with a specific focus on critical object-oriented reasoning. Our approach actively leverages a pre-trained VLM to generate high-level semantic decisions through a Chain-of-Thought (CoT) prompting mechanism, shown in Fig.~\ref{fig:problem}(c). By explicitly reasoning about critical traffic elements in complex driving scenarios, the VLM transforms multi-view image inputs into structured semantic decision priors that guide downstream RL training. This structured guidance reduces policy exploration complexity, promotes safe and coherent driving behavior, and establishes a clear interpretative link between the visual environment, key objects, and the resulting control actions.
The main contribution of this paper is list as follows:

\begin{itemize}
    \item We propose COVLM-RL, a novel end-to-end RL framework guided by VLM, centered on critical object-oriented reasoning. By introducing CoT prompting mechanism inspired by human driving cognition "Identification-Prediction-Planning", our method enables the VLM to produce structured, high-level semantic decisions from complex multi-view inputs, guiding the learning process with interpretable priors.
    \item To bridge the gap between high-level semantic reasoning and low-level control execution, we further introduce a consistency loss that enforces alignment between VLM-generated plans and RL-generated actions. This improves training stability and promotes semantically grounded decision-making.
    \item 
    We validate our framework through extensive experiments in the CARLA simulator. Results demonstrate that our method significantly improves the success rate by 30\% in trained driving environments and by 50\% in previously unseen environments, highlighting its strong generalization capability.
\end{itemize}

\section{RELATED WORK}

\subsection{End-to-End Autonomous Driving}

End-to-end autonomous driving (E2E-AD) frameworks have gained prominence as a unified alternative to traditional modular pipelines, which separately handle perception, planning, and control. By learning a direct mapping from raw sensory inputs to control actions, E2E-AD models aim to reduce error propagation between modules and improve system scalability and latency \cite{DriveAnywhere,song2024autoware}. Despite these advantages, E2E-AD systems face substantial challenges in training efficiency, generalization to unseen environments, and policy interpretability.

Supervised learning remains the dominant paradigm for training E2E-AD models \cite{chen2023e2esurvey}, requiring large-scale, diverse datasets with high-quality action labels. However, collecting such datasets in real-world traffic scenarios is expensive, time-consuming, and inherently limited in coverage, particularly for edge cases and safety-critical interactions. Imitation learning mitigates some of these costs by training policies to mimic expert demonstrations \cite{codevilla2018end,zhang2021end}, but it suffers from compounding errors, limited diversity, and poor scalability in long-horizon or interactive tasks.
To overcome these limitations, RL has emerged as a promising alternative. RL enables agents to learn optimal driving behaviors through trial-and-error interactions with the environment, removing the need for exhaustive supervision and allowing adaptation to novel situations \cite{cai2024interaction}. Simulation platforms such as CARLA \cite{Dosovitskiy17} facilitate scalable and safe RL training under diverse and configurable driving conditions. Nevertheless, vision-based RL methods still face issues related to sample inefficiency, representation learning, and decision-making transparency.

\subsection{VLM-Guided Reinforcement Learning}
In reinforcement learning, visual feature extractors are crucial for encoding raw image observations into representations suitable for decision-making. Traditional visual backbones, such as CNNs or ResNet variants \cite{10378442}, often suffer from information loss and poor semantic abstraction when dealing with high-dimensional visual inputs. The recent emergence of VLMs, such as CLIP \cite{radford2021learning}, has introduced new possibilities for multi-modal representation learning. VLMs align image features with natural language, enabling improved interpretability, semantic grounding, and generalization across diverse tasks.
\begin{figure*}
    \centering
    \includegraphics[width=18cm]{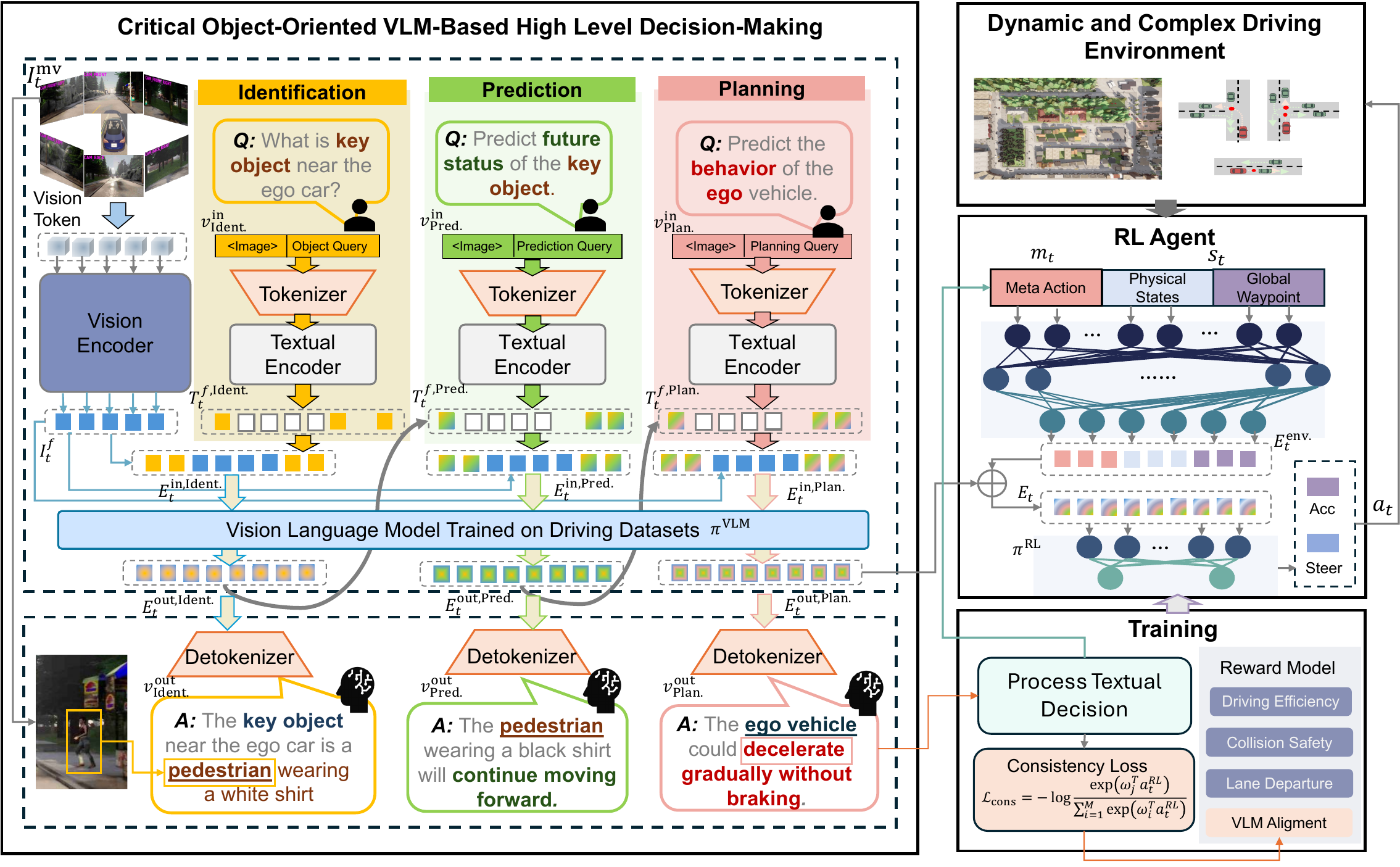}
    \caption{
    Overview of the proposed \textbf{COVLM-RL} framework. The system integrates a critical object-oriented CoT reasoning module with a pretrained VLM to generate high-level semantic guidance from multi-view camera inputs. The CoT module sequentially identifies key traffic objects, predicts their behavior, and plans an appropriate ego action in natural language. This action is parsed into both a discrete meta-action (one-hot) and a semantic embedding. The meta-action and its associated CoT features are concatenated with environment observations (physical state and global waypoint) and used as input to the RL policy. 
    During training, a consistency loss encourages alignment between the RL agent's control outputs and the VLM's semantic intent.
    }
    \label{fig:framework}
\end{figure*}
Existing research on integrating RL with VLMs can be broadly categorized into two approaches. The first involves RL from human feedback (RLHF) \cite{sun2024optimizing, wang2024reinforcement}, where reward models are learned from human preferences prior to RL deployment. The second employs task-specific reward functions without reliance on human preference data\cite{huang2024vlm, doroudian2024clip}. However, training a VLM from scratch demands large-scale, diverse, domain-adapted labeled datasets and substantial computational resources, and they do not adequately focus on critical objects, limiting their effectiveness in safety-critical driving environments.

Our approach distinguishes itself by explicitly integrating critical object-oriented reasoning via a pretrained VLM within the RL training loop. By exploiting the structured, interpretable semantic knowledge from VLMs, we significantly enhance policy robustness, training efficiency, and interpretability, particularly in complex traffic scenarios involving critical object interactions.

\section{METHODOLOGY}


Generalizing decision-making policies to diverse and safety-critical driving environments remains a core challenge in autonomous vehicle (AV) systems. In this work, we address the task of end-to-end driving control, where the agent must follow global waypoints while avoiding collisions and adhering to traffic constraints. To this end, we propose COVLM-RL, a novel framework that integrates high-level semantic reasoning from  pretrained VLM into the RL training loop. As illustrated in Fig.~\ref{fig:framework}, our method employs CoT prompting strategy to extract critical object-oriented semantic guidance from multi-view visual inputs. The VLM outputs a high-level driving intention in natural language, which is parsed into a discrete meta-actions and semantic embedding. These are combined with the vehicle's physical state and waypoint information to form the full observation input to the RL policy. During training, we introduce a consistency loss that aligns the RL agent's low-level control outputs with the VLM's semantic guidance, improving both interpretability and learning stability. To manage computational cost, the VLM is queried every $N$ steps, while the RL agent operates at a higher frequency. This design provides structured task-relevant priors, accelerates convergence, and enables robust policy learning in complex and previously unseen scenarios. 
The following sections detail our formulation of the VLM-augmented RL environment, the proposed CO reasoning module, and the training strategy used to integrate high-level VLM semantic priors and low-level RL actions.

\subsection{VLM-Augmented RL Formulation}

We formulate the decision-making problem as a VLM-augmented Markov Decision Process (MDP), defined by the tuple $\langle \mathcal{S}, \mathcal{A}, \mathcal{T}, \mathcal{R}, \gamma \rangle$. Here, $\mathcal{S}$ is the state space, which we decompose into the sensory observation space $\mathcal{O}$ and the meta-action space $\mathcal{M}$ generated by the VLM, i.e., $\mathcal{S} = \mathcal{O} \times \mathcal{M}$. The action space $\mathcal{A}$ includes the ego vehicle's low-level control signals (e.g., steering and acceleration). The transition function $\mathcal{T}: \mathcal{S} \times \mathcal{A} \times \mathcal{S} \rightarrow [0, 1]$ models environment dynamics, and the reward function $\mathcal{R}: \mathcal{S} \times \mathcal{A} \rightarrow \mathbb{R}$ measures task performance.

To generate semantic guidance, a pretrained VLM is queried via a prompting policy $\pi^{\text{VLM}}: \mathcal{I} \times \mathcal{V}^{\text{in}} \rightarrow \mathcal{V}^{\text{out}}$, where $I^{\text{mv}}_t \in \mathcal{I}$ denotes the multi-view camera input at time $t$, and $v^{\text{in}} \in \mathcal{V}^{\text{in}}, v^{\text{out}} \in \mathcal{V}^{\text{out}}$ are the input and output tokens. A parsing module maps the VLM's natural language output into a vectorized meta-action: $\mathcal{V}^{\text{out}} \rightarrow \mathcal{M}$, bridging symbolic CoT reasoning with motion planning module of the RL agent.

Our policy $\pi^{\theta}$ is implemented with two components: a feature extractor that encodes environmental observations $E_t^{\text{env}}$ and the VLM-generated planning embedding $E_t^{\text{out,Plan}}$, and a multilayer perceptron (MLP) that maps the concatenated feature vector $E_t = [E_t^{\text{env}}, E_t^{\text{out,Plan}}]$ to a control action $a_t \in \mathcal{A}$.
The agent's objective is to maximize the expected cumulative discounted reward:
\begin{equation}
    \mathbb{E}\left[\sum_{t=1}^{\infty} \gamma^t \mathcal{R}(s_t, a_t) \ \middle| \ a_t \sim \pi^{\theta}(a_t \mid E_t^{\text{env}}, E_t^{\text{out,Plan}}) \right]
\end{equation}
where $a_t$ is sampled from the policy $\pi^{\theta}(a_t \mid E_t)$. Formally, the optimal policy is defined as:
\begin{equation}
    a_t^* = \arg\max_{a} \mathbb{E}\left[\sum_{t=1}^{\infty} \gamma^t \mathcal{R}(s_t, a_t) \mid \pi^{\theta}(a_t \mid E_t)\right]
\end{equation}

This formulation enables the RL agent to leverage high-level semantic priors generated by the VLM, thereby improving training efficiency, sample efficiency, and adaptability in complex and dynamic driving environments.




\subsubsection{Driving Environment States}

To operate effectively across diverse scenarios such as urban intersections, turns, and lane merges, the state representation must include both low-level physical observations and high-level semantic guidance. We define the observation at time step $t$ as:
\begin{equation}
    s_t = [o_t, m_t] = [\tau_t, s_t^{\text{phys}}, m_t] = [\tau_t, \delta_t, \alpha_t, v_t, m_t]
\end{equation}
Here, $\tau_t$ represents the reference waypoints obtained from the global navigation map, consisting of a sequence of target positions $(x_t^i, y_t^i)$. The physical state $s_t^{\text{phys}}$ includes the current steering angle $\delta_t$, throttle value $\alpha_t$, and speed $v_t$ of the ego vehicle.

The meta-action $m_t$ is a high-level driving intent generated by the VLM through Chain-of-Thought reasoning (e.g., "turn left" or "decelerate"), and is encoded as a one-hot vector to be processed by the RL policy. This semantic input grounds the policy in task-relevant abstractions, improving both interpretability and learning efficiency.

\begin{table}
\caption{Description of Meta-Actions}
\label{tab:action description}
\tiny
\resizebox{\columnwidth}{!}{%
\begin{tabular}{ll}
\hline
\textbf{Action} & \textbf{Description} \\
\hline
SLOW   & Reduce speed or maintain a lower velocity \\
FAST   & Accelerate or maintain a higher speed     \\
LEFT   & Steer or move to the left                 \\
RIGHT  & Steer or move to the right                \\
IDEL   & Maintain the current state with minimal control \\
\hline
\end{tabular}%
}
\end{table}



\subsubsection{Action Space}

To enable end-to-end control, the RL agent directly outputs executable vehicle control commands. At each time step $t$, the action $a_t$ consists of continuous control variables for longitudinal and lateral motion:
\begin{equation}
    a_t = [\alpha_t, \delta_t]
\end{equation}
where $\alpha_t$ is the throttle (acceleration) and $\delta_t$ is the steering angle. These signals are applied to the vehicle dynamics model to control the ego vehicle, enabling seamless interaction between the learned policy and the environment.



\subsubsection{Reward Function}

The reward function is designed to guide the RL agent's behavior in dynamic and safety-critical driving environments. It integrates essential driving objectives-including safety, efficiency, and adherence to road rules-into a scalar signal used for policy optimization. Formally, the total reward at each timestep is computed as:
\begin{equation}
    r_t = r_t^{\text{collision}} + r_t^{\text{eff}} + r_t^{\text{lane}}
\end{equation}
Here, $r_t^{\text{collision}}$ penalizes collisions with other vehicles or obstacles to promote safe driving behavior. The term $r_t^{\text{eff}}$ encourages the agent to maintain appropriate speed and avoid unnecessary delays, thereby improving travel efficiency. The third term, $r_t^{\text{lane}}$, penalizes lateral deviation from the center of the lane, reinforcing proper lane-keeping and minimizing off-road driving.
Together, these components provide a well-balanced reward structure that supports the learning of robust and generalizable policies across diverse urban scenarios. All parameter settings for the reward terms are detailed in Table~\ref{tab:reward_params}.

\begin{table}
\caption{Reward Function Parameters}
\label{tab:reward_params}
\centering
\small
\begin{tabular}{ll}
\toprule
\textbf{Parameter} & \textbf{Value} \\
\midrule
Min speed & 0.0 \\
Max speed & 28.8 \\
Target speed & 25.0 \\
Max distance & 4.0 \\
Max std center lane & 0.4 \\
Max angle enter lane & 90 \\
Penalty reward & -10 \\
\bottomrule
\end{tabular}
\end{table}

\subsection{Critical Object-Oriented VLM-Based Decision Making}
 

This module is responsible for transforming multi-view visual observations into semantically meaningful high-level driving decisions using a CoT prompting strategy. Specifically, we adopt Mini-InternVL \cite{chen2024internvl}, an open-source pretrained VLM trained on the DriveLM dataset \cite{sima2023drivelm}. Mini-InternVL is selected for its lightweight design and generalization capacity, making it well-suited for onboard vehicle deployment. Below, we describe the VLM architecture, the construction of multi-view image inputs, and the CoT prompting mechanism used to guide object-centric reasoning. While our implementation is based on Mini-InternVL, the overall framework is compatible with other foundation VLMs that support structured visual-textual reasoning.

\subsubsection{VLM Model Architecture}

The selected model, Mini-InternVL, is a lightweight VLM designed to align a large-scale vision encoder with an existing LLM, making it suitable for deployment in resource-constrained autonomous vehicles. Mini-InternVL comprises three main components: InternViT as visual encoder, a MLP projector, and Qwen2 \cite{yang2024qwen2} as Language model. Given an input image $I \in \mathbb{R}^{H \times W \times 3}$, the visual encoder extracts a feature map $I^f \in \mathbb{R}^{N \times D}$, which is then projected into the language embedding space via the MLP connector. This architecture enables the model to jointly leverage visual and linguistic modalities for downstream decision-making tasks.

\subsubsection{Multi-View Visual Encoding}



Unlike monocular camera setups, multi-view visual input enables the construction of a unified spatial representation of the ego vehicle's surroundings, thereby eliminating blind spots and improving scene understanding. At each time step $t$, the set of multi-view images is defined as:
\begin{equation}
    I_t^{\text{mv}} = \{ I_t^{\text{front,left}}, I_t^{\text{front}}, I_t^{\text{front,right}}, I_t^{\text{rear,left}}, I_t^{\text{rear}}, I_t^{\text{rear,right}} \}
\end{equation}
corresponding to six camera perspectives arranged around the ego vehicle.
Each image is first resized to a resolution of $2H \times H$ and then concatenated in a fixed spatial sequence to form a panoramic composite with a final resolution of $3 \times 2H \times 2H$, i.e., $I_t' \in \mathbb{R}^{3 \cdot 2H \times 2H \times 3}$. To further enhance the model's ability to capture fine-grained spatial cues, we adopt a dynamic resolution scheme following \cite{li2024driving}. Specifically, the combined image is divided into thirteen $H \times H$ patches and transformed into 256 visual tokens using the visual encoder.

\subsubsection{Critical Object-Oriented Chain-of-Thought}

To mitigate hallucinations commonly observed in foundation models, we propose a critical object-oriented (CO) CoT prompting strategy tailored for safety-critical autonomous driving scenarios \cite{liao2025cot}. Inspired by the cognitive process of human drivers, our method decomposes high-level reasoning into a structured sequence: identification, prediction, and planning.


At each update step $t$, the model first identifies the most critical object in the scene (e.g., a cut-in vehicle or a pedestrian crossing suddenly), then predicts its future behavior, and finally plans the ego vehicle's response based on prior outputs. This structured reasoning process significantly improves the reliability and interpretability of the VLM's decisions.
We denote the full dialogue process at time $t$ as:
\begin{equation}
    \mathcal{D}_t = \{v^{\text{in}}_{\text{Ident.}}, v^{\text{out}}_{\text{Ident.},t}, v^{\text{in}}_{\text{Pred.}}, v^{\text{out}}_{\text{Pred.},t},v^{\text{in}}_{\text{Plan.}}, v^{\text{out}}_{\text{Plan.},t}\}
\end{equation}



where $v^{\text{in}}_{\text{Ident.}}$, $v^{\text{in}}_{\text{Pred.}}$, and $v^{\text{in}}_{\text{Plan.}}$ denote the textual prompts for identification, prediction, and planning, respectively.
Given a preprocessed composite image $I'$, the model performs the three-step reasoning as follows:

(1) Identification: The VLM processes $I'$ with the identification prompt to detect the most salient object:
\begin{equation}
    v^{\text{out}}_{\text{Ident.},t}, E^{\text{out}}_{\text{Ident.},t} = \textbf{VLM}(I', v^{\text{in}}_{\text{Ident.}})
\end{equation}

(2) Prediction: Conditioned on the identification output, the model predicts the future behavior of the object:
\begin{equation}
    v^{\text{out}}_{\text{Pred.},t}, E^{\text{out}}_{\text{Pred.},t} = \textbf{VLM}(I', v^{\text{in}}_{\text{Pred.}} \mid E^{\text{out}}_{\text{Ident.},t})
\end{equation}

(3) Planning: Finally, the model plans a high-level driving action for the ego vehicle:
\begin{equation}
    v^{\text{out}}_{\text{Plan.},t}, E^{\text{out}}_{\text{Plan.},t} = \textbf{VLM}(I', v^{\text{in}}_{\text{Plan.}} \mid E^{\text{out}}_{\text{Ident.},t}, E^{\text{out}}_{\text{Pred.},t})
\end{equation}

\begin{figure}
    \centering
    \includegraphics[width=\linewidth]{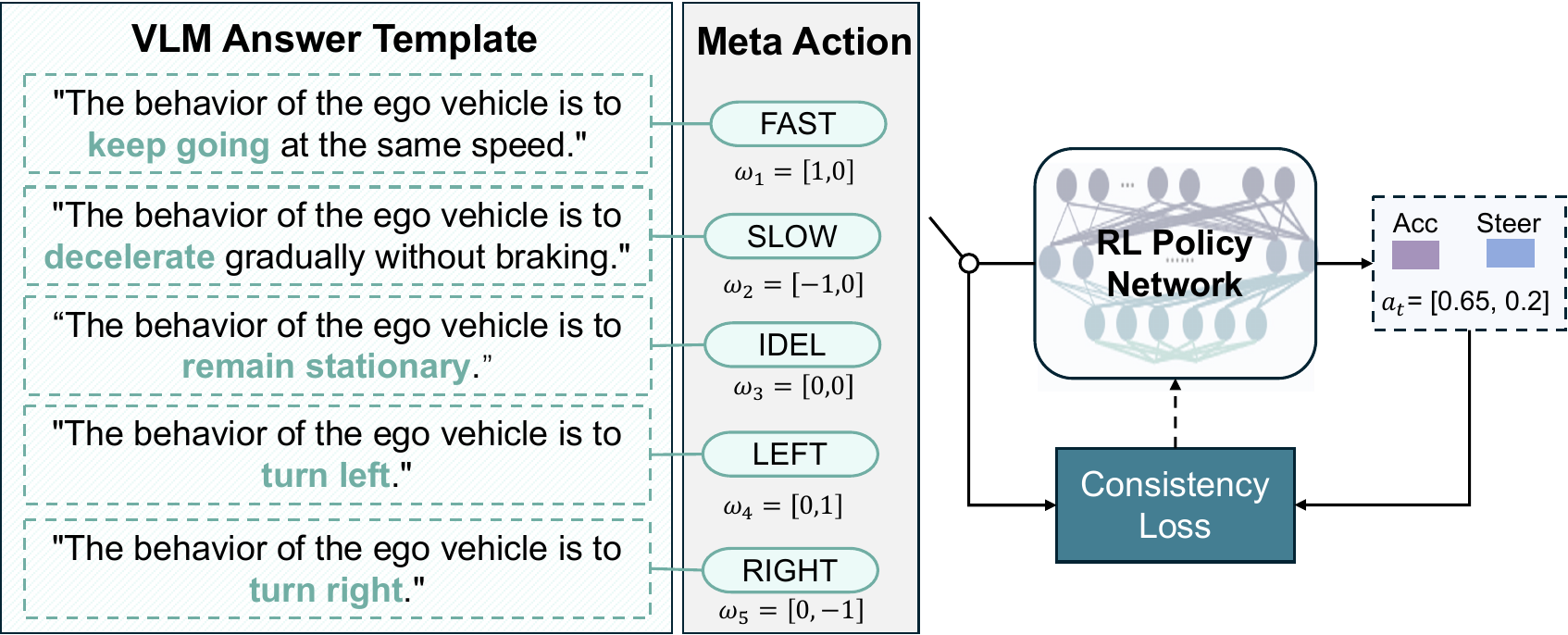}
    \caption{Overview of the parsing process for high-level semantic decisions produced by the VLM.}
    \label{fig:parse}
\end{figure}

This critical object-oriented CoT prompting strategy encourages interpretable, step-by-step reasoning grounded in both the visual context and the dynamic behavior of critical objects. To balance semantic guidance with computational efficiency, the VLM is queried once every $N$ steps, and its outputs are reused across intermediate RL steps. This temporal decoupling introduces a trade-off between guidance granularity and inference time, but the high-level decisions remain effective in providing consistent priors that support robust policy learning. The full pipeline for parsing textual decisions is illustrated in Fig.~\ref{fig:parse}.

\subsection{VLM-Guided RL Training Strategy}


Compared to traditional DRL policy, integrating a VLM into the RL training loop introduces high-level semantic reasoning that can substantially improve learning efficiency and policy generalization. By providing structured, task-relevant priors, the VLM enables the RL agent to explore more intelligently, especially in safety-critical scenarios. However, this integration presents new challenges. In particular, the augmented observation space-comprising both low-level sensory inputs and high-dimensional VLM outputs-can introduce conflicting learning signals. 
To address these issues, we propose a VLM-guided training strategy that incorporates a semantic consistency loss. This loss aligns the agent's low-level control actions with the VLM's high-level decisions, serving both as a semantic regularizer and a bridge across modalities. The following subsections detail the alignment mechanism and the formulation of the consistency objective.

\subsubsection{RL with VLM-Guided Semantic Alignment}
The increased dimensionality of the observation space, resulting from the integration of VLM outputs, can introduce conflicting learning signals and amplify training instability, particularly due to the delayed and sparse nature of RL rewards and the long-term dependencies inherent in sequential decision-making tasks. To address this, we introduce a consistency loss term into the overall training objective to encourage alignment between the RL policy and the semantic guidance provided by the VLM. This facilitates the agent's ability to associate the success or failure of a task with the high-level decisions of the VLM, thereby improving the interpretability and effectiveness of policy learning.

\begin{equation} 
\mathcal{L}^{\text{total}} = \mathcal{L}^{\text{RL}} + \lambda \mathcal{L}^{\text{Cons}} 
\end{equation}

Here, $\mathcal{L}^{\text{RL}}$ denotes the standard RL loss aimed at maximizing expected cumulative rewards, and $\lambda$ is a hyperparameter that balances the contribution of the consistency regularization term $\mathcal{L}^{\text{Cons}}$.

\begin{figure*}[h]
\centering
    \includegraphics[width=17.5cm]{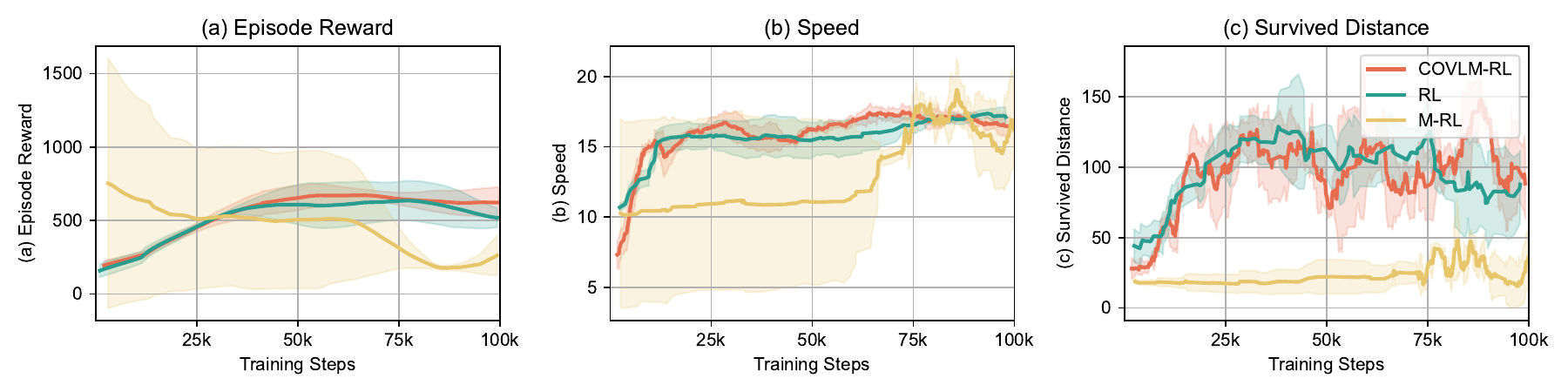}
    \caption{Training curves of different methods:(a) Episode Reward (b) Speed (c) Survived distance.}
    \label{fig:training_curves}
\end{figure*}

\subsubsection{Consistency Loss for Semantic-Action Alignment}


To bridge the modality gap between the VLM's textual outputs and the RL agent's continuous control actions, we introduce a conversion module that transforms the natural language plan $v^{\text{out}}_{\text{Plan.},t}$ into two aligned forms: a discrete meta-action label $m_t$ and its corresponding semantic embedding $\omega_t$:

\begin{equation}
    m_t, \omega_t = \textbf{Convert}(v^{\text{out}}_{\text{Plan.},t})
\end{equation}

Here, $m_t$ is a one-hot vector representing the meta-action class (e.g., "SLOW", "LEFT"), and $\omega_t \in \mathbb{R}^d$ is a dense semantic embedding used for regularization.
To further enforce alignment between the VLM's semantic guidance and the RL agent's actions, we define a contrastive consistency loss that penalizes mismatches between the selected action $a_t$ and the target semantic embedding $\omega_j$:

\begin{equation}
    \mathcal{L}^{\text{Cons}}=-\log \frac{\exp (\omega^T_ja_t)}{(\omega^T_ia_t)}
\end{equation}


This loss encourages the policy's output $a_t$ to be maximally similar to the semantically correct VLM embedding $\omega_j$, thereby enforcing semantic consistency in the learned control behavior.

\begin{table}[ht]
\centering
\scriptsize
\caption{Reinforcement Learning Algorithm Parameters}
\label{tab:algorithm_params}
\begin{tabular}{ll}
\toprule
\textbf{Parameter} & \textbf{Value} \\
\midrule
Learning  rate & $1\text{e}{-6}$ \\
Gamma & 0.98 \\
GAE(${\lambda}$) & 0.95 \\
Clip range & 0.2 \\
Entropy regularization coefficient & 0.05 \\
n\_epochs & 10 \\
n\_steps & 1024 \\
Policy\_kwargs & ReLU, [500, 300], 256-dim extractor \\

\bottomrule
\end{tabular}
\end{table}

\section{EXPERIMENTS}

We evaluate the effectiveness of the proposed COVLM-RL framework through extensive experiments conducted in the CARLA simulator, a high-fidelity urban driving environment widely adopted in autonomous driving research.

\subsection{Setups}

We train all models in Town02, a moderately complex urban environment characterized by T-junctions, parks, coniferous trees, and mixed residential-commercial zones. To evaluate generalization capability, we test the trained models in both Town02 (seen environment) and a previously unseen environment, Town03. Town03 features a significantly more complex road network, including roundabouts, underpasses, overpasses, elevated metro tracks, and construction zones, closely resembling a downtown urban setting.
The ego vehicle is equipped with multi-view semantic segmentation cameras and receives high-level waypoint instructions for route navigation. The action space consists of continuous control commands for steering and acceleration.

We compare our proposed \textbf{COVLM-RL} framework against two baselines. The first is a standard \textbf{RL} agent trained using Proximal Policy Optimization (PPO), which receives segmentation image inputs and global waypoints but does not incorporate semantic priors or high-level planning. The second is a maneuver-enhanced variant (\textbf{M-RL}), which incorporates additional behavior priors derived from map topology (e.g., upcoming turns, intersection layouts) to introduce basic maneuver awareness during policy learning.
In contrast, our COVLM-RL method leverages a pretrained VLM that performs object-centric CoT reasoning to generate high-level semantic driving intentions. These intentions are used to guide the RL agent via a consistency loss that aligns semantic guidance with the learned control policy.

All models are trained under identical configurations, using the same number of episodes, network architectures, and hyperparameters, as summarized in Table~\ref{tab:algorithm_params} and Table~\ref{tab:reward_params}.

\begin{figure}[h]
    \centering
    \includegraphics[width=\linewidth]{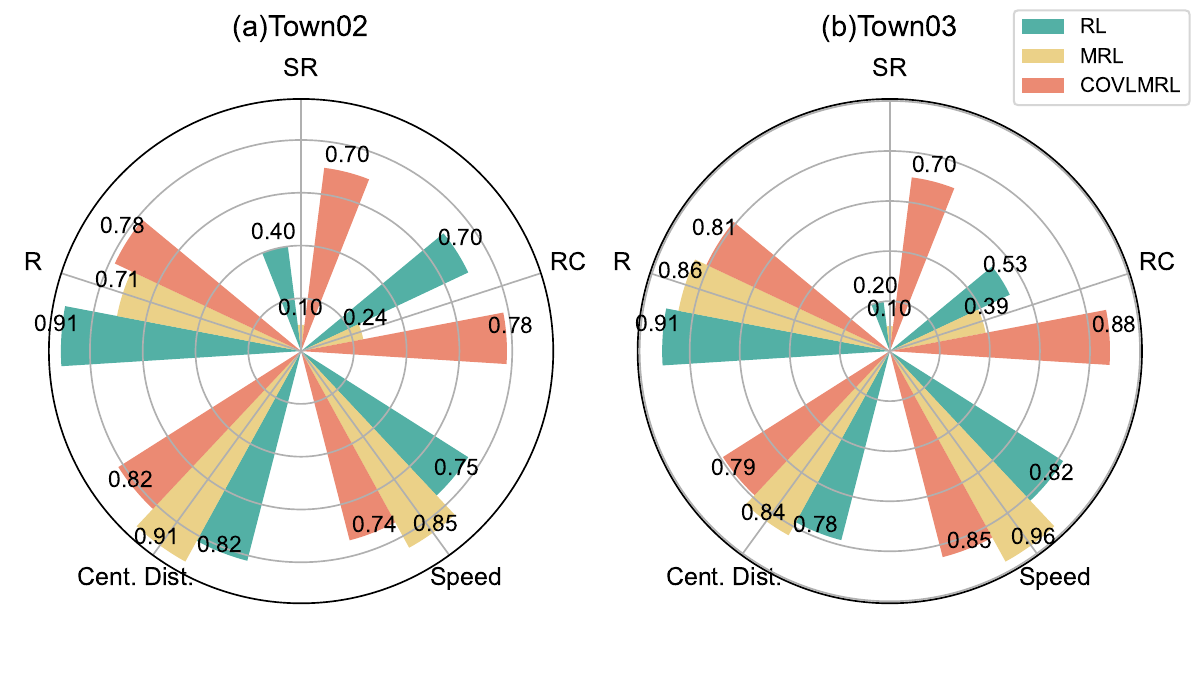}
    \caption{Evaluation Metrics of different algorithms: (a) Trained scenarios (b) Unseen scenarios.}
    \label{fig:eval}
\end{figure}

\begin{table*}[htbp]
\centering
\caption{Quantitative Performance Comparison on Trained scenarios.}
\label{tab:eval_trained_scenarios}
\resizebox{\textwidth}{!}{%
\begin{tabular}{lcccccccccc}
\toprule
\textbf{Method} & \textbf{SR~(↑)} & \textbf{TD~(↑)} & \textbf{AD~(↑)} & \textbf{RC~(↑)} & \textbf{SM~(↑)} & \textbf{SS~(↓)} & \textbf{CDM~(↓)} & \textbf{CDS~(↓)} & \textbf{RM~(↑)} & \textbf{RS~(↓)} \\
\midrule
RL      & 0.40 & 1447.04 & 144.70 & 0.70 & 21.48 & 4.48 & 0.69 & 0.54 & 5.03 & 1.64 \\
M-RL    & 0.10 & 201.43 & 20.14 & 0.24 & 18.84 & 9.07 & 0.33 & 0.31 & 3.95 & 2.17 \\
\textbf{COVLM-RL} & \textbf{0.70} & \textbf{1458.95} & \textbf{145.89} & \textbf{0.78} & \textbf{18.74} & \textbf{3.34} & \textbf{0.71} & \textbf{0.63} & \textbf{4.29} & \textbf{1.20} \\
\bottomrule
\end{tabular}
}
\end{table*}

\begin{table*}[ht]
\centering
\caption{Quantitative Performance Comparison on Evaluation Unseen scenarios.}
\resizebox{\textwidth}{!}{%
\begin{tabular}{lcccccccccc}
\toprule
\textbf{Method} & \textbf{SR~(↑)} & \textbf{TD~(↑)} & \textbf{AD~(↑)} & \textbf{RC~(↑)} & \textbf{SM~(↑)} & \textbf{SS~(↓)} & \textbf{CDM~(↓)} & \textbf{CDS~(↓)} & \textbf{RM~(↑)} & \textbf{RS~(↓)} \\
\midrule
RL      & 0.20 & 2070.93 & 207.10 & 0.53 & 20.62 & 4.22 & 0.85 & 0.49 & 4.98 & 1.54 \\
M-RL    & 0.10 & 1529.71 & 152.97 & 0.39 & 24.24 & 4.50 & 0.63 & 0.55 & 4.76 & 2.04 \\
\textbf{COVLM-RL}  & \textbf{0.70} & \textbf{3468.47} & \textbf{346.96} & \textbf{0.88} & \textbf{21.38} & \textbf{3.25} & \textbf{0.82} & \textbf{0.64} & \textbf{4.41} & \textbf{1.33} \\
\bottomrule
\end{tabular}
\label{tab:ppo_comparison}}
\end{table*}

\subsection{Training Results}


To assess the effectiveness of the proposed COVLM-RL framework, we compare its training performance against the RL and M-RL baselines. As illustrated in Fig.~\ref{fig:training_curves}, COVLM-RL consistently achieves superior results across key metrics, including episode reward, average speed, and survived distance.
Interestingly, the M-RL baseline underperforms relative to the vanilla RL agent, suggesting that incorporating maneuver priors without sufficient semantic grounding may introduce biases or constrain policy exploration. In contrast, COVLM-RL leverages structured, object-centric semantic guidance produced by a pretrained VLM via CoT reasoning. These high-level decisions, when aligned with low-level control through the proposed consistency loss, enable more efficient and stable policy optimization. The results indicate that simple heuristic augmentation is insufficient-high-quality, interpretable semantic priors are critical for robust RL performance in complex driving environments.

\subsection{Evaluation Results}
To evaluate agent performance, we report a comprehensive set of driving metrics across both trained and unseen scenarios, visualized as Fig.~\ref{fig:eval}. \text{Success Rate (SR)} measures the proportion of episodes in which the agent reaches the destination without failure. \text{Traveled Distance (TD)} and \text{Average Distance (AD)} represent the total and per-episode distances traveled, respectively, indicating driving efficiency. \text{Route Completion (RC)} quantifies the percentage of the planned route completed, serving as a proxy for task fulfillment. \text{Speed Mean (SM)} and \text{Speed Std (SS)} reflect the average and variability in driving speed, capturing motion smoothness. \text{Centerline Deviation Mean (CDM)} and \text{Centerline Deviation Std (CDS)} evaluate the mean and variance in lateral displacement from the lane center, indicating lane-keeping accuracy. \text{Reward Mean (RM)} and \text{Reward Std (RS)} capture the overall quality and consistency of policy behavior, respectively.

As shown in Table \ref{tab:eval_trained_scenarios}, COVLM-RL consistently outperforms both RL and M-RL across a broad range of metrics in the trained environments. It achieves the highest success rate 0.70, near-perfect route completion  0.78, and robust reward performance, reflecting both effective learning and stable policy behavior. Additionally, COVLM-RL exhibits low centerline deviation with deviation distance 0.71m, indicating accurate lane-following capability. In contrast, the M-RL baseline performs significantly worse than the standard RL agent in most dimensions, with particularly low success rate 0.10, reduced travel distance 201.43m, and high reward variance 2.17. These results suggest that the naive integration of maneuver priors without sufficient semantic grounding may restrict exploration and hinder policy optimization. By contrast, COVLM-RL benefits from object-centric semantic decisions generated by the VLM, which are better aligned with the task goals and effectively regularize the policy learning process.

As shown in Table \ref{tab:ppo_comparison}, COVLM-RL demonstrates strong generalization capabilities in previously unseen environments. It achieves a success rate of 70\%, more than three times that of the RL baseline, and the longest average distance traveled, which is 3468.47m, alongside the highest route completion score, 0.88. Despite increased scenario complexity, COVLM-RL maintains accurate lane adherence, with deviation distance 0.82m, stable and high reward returns, evidencing its robustness and consistency. In contrast, both RL and M-RL suffer performance degradation, with M-RL again exhibiting the poorest results, likely due to over-reliance on rigid heuristic priors. These findings highlight the advantage of vision-language semantic reasoning in producing robust, adaptable, and interpretable driving policies capable of generalizing to out-of-distribution scenarios.

\section{CONCLUSIONS}
In this paper, we presented COVLM-RL, a novel VLM-guided RL framework that explicitly leverages critical object-oriented reasoning to enhance the generalization, interpretability, and safety of autonomous driving policies. By integrating a pretrained VLM with a CoT prompting strategy, our approach transforms multi-view visual observations into high-level semantic decision priors, significantly simplifying the exploration space and improving policy robustness. Furthermore, we introduced a consistency loss to effectively align high-level reasoning with low-level control and encourage semantic-action coherence during training. This semantic alignment improves learning stability and grounds decision-making in interpretable object-centric cues.

Extensive experiments conducted in the CARLA simulator, across both seen and unseen urban environments, demonstrate that COVLM-RL consistently outperforms conventional RL and maneuver-enhanced RL baselines in terms of success rate, driving efficiency, lane-keeping accuracy, and reward stability. In particular, COVLM-RL achieves superior generalization to previously unseen environments, highlighting the efficacy of incorporating critical object-centric semantic reasoning into the RL training loop. Future work will focus on extending the proposed framework to real-world driving datasets and exploring adaptive prompting strategies for further improving reasoning robustness under highly dynamic traffic scenarios.







\bibliography{ref}
\bibliographystyle{IEEEtran}


\end{document}